\documentclass[lettersize,journal]{IEEEtran}
\usepackage{amsmath,amsfonts}
\usepackage{algorithm}
\usepackage{algorithmicx}
\usepackage{algpseudocode}
\usepackage{array}
\usepackage[caption=false,font=normalsize,labelfont=sf,textfont=sf]{subfig}
\usepackage{textcomp}
\usepackage{stfloats}
\usepackage{url}
\usepackage{verbatim}
\usepackage{graphicx}
\usepackage[
        numbers,square,
        ]{natbib}

\usepackage[table,xcdraw,dvipsnames]{xcolor}
\usepackage{relsize}
\definecolor{cvprblue}{rgb}{0.21,0.49,0.74}
\usepackage[
        breaklinks,
        colorlinks,
        citecolor=blue,
        linkcolor=blue,
        ]{hyperref}

\usepackage{multirow}
\usepackage{microtype}
\usepackage{booktabs} 
\usepackage{makecell}
\usepackage{bbding}
\usepackage{tcolorbox}
\usepackage{mathtools}
\usepackage{amsthm}
\usepackage{makecell}

\begin{document}

\title{\textsc{AgentsCoDriver}: Large Language Model Empowered Collaborative Driving with Lifelong Learning
\thanks{This work was supported in part by the Hong Kong Innovation and Technology Commission under InnoHK Project CIMDA, by the Hong Kong SAR Government under the Global STEM Professorship, and by the Hong Kong Jockey Club under JC STEM Lab of Smart City.}
}

\author{Senkang Hu, Zhengru Fang, Zihan Fang,~\IEEEmembership{Graduate Student Member,~IEEE,} Yiqin Deng,~\IEEEmembership{Member,~IEEE,} \\Xianhao Chen,~\IEEEmembership{Member,~IEEE,} Yuguang Fang,~\IEEEmembership{Fellow,~IEEE} 
\thanks{Senkang Hu, Zhengru Fang, Zihan Fang, and Yuguang Fang are with the Department of Computer
Science, City University of Hong Kong, Kowloon, Hong Kong. (e-mail: \texttt{\{senkang.forest, zhefang4-c\}@my.cityu.edu.hk, \{zihanfang3, my.Fang\}@cityu.edu.hk})}
\thanks{Yiqin Deng is with the School of Control Science and Engineering, Shandong University, Jinan, China. (e-mail: \texttt{yiqin.deng@email.sdu.edu.cn})}
\thanks{Xianhao Chen is with the Department of Electrical and Electronic Engineering,
The University of Hong Kong, Pok Fu Lam, Hong Kong. (e-mail: \texttt{xchen@eee.hku.hk})}

}


\maketitle
\begin{abstract}
Connected and autonomous driving has gone through rapid development lately. However, current autonomous driving systems, which are primarily based on data-driven approaches, exhibit significant deficiencies in interpretability, generalization, and continuing learning capabilities. In addition, single-vehicle autonomous driving systems lack the ability of collaboration and negotiation with other vehicles, which is crucial for driving safety and efficiency. In order to effectively address these issues, we leverage large language models (LLMs) to develop  a novel framework, called \textsc{AgentsCoDriver}, to enable multiple vehicles to conduct collaborative driving. \textsc{AgentsCoDriver} consists of five modules: observation module, reasoning engine, cognitive memory module, reinforcement reflection module, and communication module. It can accumulate knowledge, lessons, and experience over time by continuously interacting with the driving environment, thereby making it possible to achieve lifelong learning. In addition, by leveraging the communication module, different agents can exchange information and realize negotiation and collaboration in complex driving  environments. Extensive experiments are conducted and show the superiority of \textsc{{AgentsCoDriver}} to existing approaches.
\end{abstract}


\section{Introdcution}

\begin{figure}[t]
    \centering
    \includegraphics[width=1\linewidth]{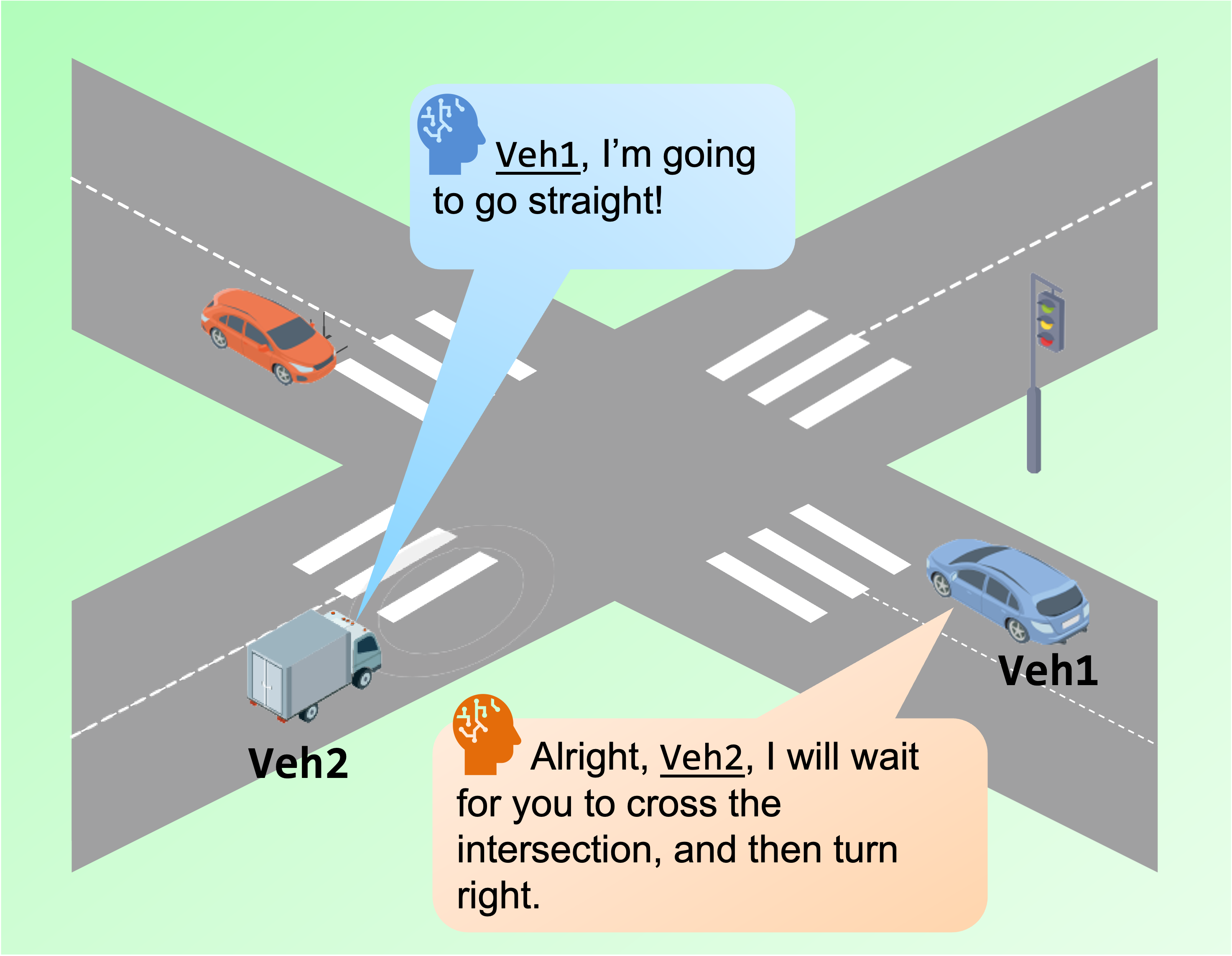}
    \caption{The Scenario of Multi-Vehicle Negotiation.}
    \label{fig:multi-vehicle-collaborative-driving}
\end{figure}


With rapid development of artificial intelligence (AI) and deep learning (DL), autonomous driving has attracted great attention from both industries and academia and tremendous progress have been made lately. However, most current autonomous driving systems are based on data-driven approaches, which faces several severe limitations. Firstly, informational interpretability is crucial for humans to underscore the safety and build trust in autonomous driving systems. 
However, the conventional approaches lack interpretability, which is a persistent problem for existing autonomous driving systems. Since the conventional methods are based on black-box deep neural networks, so it is not viable to understand the process of decision-making, reasoning, and logic thinking behind a decision.  
Another limitation is the lack of the ability of continuous learning. Since the data-driven approaches just learn from the data during the training process, and the learned knowledge is fixed and cannot be updated and adapted accordingly unless retraining the model with new incoming data, which make it difficult to learn new knowledge and adapt to new environments. In addition, traditional autonomous driving systems are based on single-vehicle decision making, which has several inherent limitations to handle complicated driving situations for a safe and reliable autonomous driving system. For example, occlusion presents a significant challenge that cannot be resolved by a single-vehicle system alone, potentially resulting in severe accidents. Moreover, the sensing range of such a system is constrained by the limitations of its sensors, potentially missing critical information. Furthermore, the single-vehicle autonomous driving systems lack the ability of collaboration and negotiation with other vehicles, which is crucial for the driving safety and efficiency in future autonomous driving systems. This leads to connected and autonomous driving (CAD), or generically called collaborative driving, which is based on a multi-vehicle system setting where an ego vehicle collaborates and negotiates with other nearby vehicles when conducting actions such as overtaking, lane changing, and left or right turning. For example, as shown in Fig. \ref{fig:multi-vehicle-collaborative-driving}, when \texttt{veh1} and \texttt{veh2} are approaching the intersection, \texttt{veh1} intends to turn right while \texttt{veh2} intends to go straight. In this case, \texttt{veh1} needs to negotiate with \texttt{veh2} to determine the order of passing through the intersection, which cannot be done without a reliable communications network.

In order to overcome the aforementioned limitations, a powerful framework is needed, and it is highly challenging to design such a framework by data-driven approaches alone. Recently, large language models (LLMs) emerge as a revolutionary technique in AI to handle more complicated tasks. LLMs are the early versions of artificial general intelligence (AGI) \cite{bubeck2023sparks},  which can think like a human, reason about the new scenarios by its own commonsense knowledge, accumulate knowledge from past experiences, and attempt interpret the decision process. These abilities make LLMs a promising technique to overcome the aforementioned limitations and build a powerful framework for collaborative driving. 

Thus, in this paper, we leverage LLMs to build a novel framework, named  \textsc{AgentsCoDriver}, which can make driving decisions and communicate with other vehicles to enable collaborative driving capability with lifelong learning. \textsc{AgentsCoDriver} consists of five parts, namely,  an observation module to encode the current traffic scenario, a cognitive memory module to store and recall different types of memories, a reasoning engine to make decisions based on the information provided by other modules, an reinforcement reflection module to reflect the output and  decision-making process, and a communication module to communicate with other vehicles to negotiate and collaborate. The main contributions of this paper are summarized as follows.
\begin{itemize}
    \item We propose a novel framework, \textsc{AgentsCoDriver}, which consists of observation, cognitive memory, reasoning engine, iterative reinforcement reflection, and communication modules, to enable multi-vehicle collaborative driving.
    \item We conduct extensive experiments to verify the effectiveness of our proposed framework, which shows its capabilities of lifelong learning, and collaboration and negotiation among multiple vehicles. 
    \item To the best of our knowledge, this is the first paper to leverage LLMs to enable multi-vehicle collaborative driving.
\end{itemize}

The remainder of this paper is organized as follows. In Sec. \ref{sec:system_formulation}, we introduce the system formulation of multi-vehicle collaborative driving. In Sec. \ref{sec:method}, we present the details modules of \textsc{AgentsCoDriver}. In Sec. \ref{sec:experiments}, we conduct extensive experiments to verify the effectiveness of \textsc{AgentsCoDriver}. In Sec. \ref{sec:limitations}, we discuss the limitations of \textsc{AgentsCoDriver} and the future works. Finally, in Sec. \ref{sec:conclusion}, we conclude this paper. 

\section{Related Work}
\label{sec:related_work}
\subsection{Development of Large Language Models}

Recent development of Large Language Models (LLMs), such as GPT-3 \cite{brown2020language}, PaLM \cite{chowdhery2023palm}, and GPT-4 \cite{sanderson2023gpt}, has revolutionized the field of artificial intelligence (AI). These models, characterized by their immense parameter size and extensive datasets for pre-training, have demonstrated remarkable capabilities in text generation and comprehension. A key feature distinguishing LLMs from their predecessors is their emergent abilities, such as in-context learning, instruction following, and reasoning with chain-of-thought, which are instrumental in their advanced performance \cite{,wang2022self,zhang2022automatic,linPushingLargeLanguage2023,lin2024efficient,fang2024automated}. Moreover, the alignment of LLMs with human objectives, such as truthfulness and the elimination of biases, has been a focal point of research \cite{wolf2023fundamental,song2023preference}. This is crucial because biases in LLMs, reflecting those of their training data sources, can significantly impact their output \cite{blodgett2020language,sheng2021societal}. Efforts to generate opinions that find consensus across diverse groups \cite{bakker2022fine} or to design prompts to mitigate biases \cite{liu2023pre} exemplify this direction. LLMs, like ChatGPT and GPT-4, have opened new avenues toward Artificial General Intelligence (AGI), showcasing human-like intelligence \cite{zhao2023survey,zhu2023minigpt}. Their application extends beyond text generation to domains like robotic tasks, such as design, navigation, and planning \cite{shah2023navigation,dorbala2023can}. The integration of LLMs in planning, particularly in dynamic, multi-agent environments, has shown promise \cite{huang2022inner}. These models can quickly understand and navigate unfamiliar environments, a task traditionally challenging for standard planning methods. This capability is particularly effective for autonomous driving, where rapid decision-making in varying contexts is critical. 

\subsection{Large Language Models for Autonomous Driving}

In autonomous driving, significant advances in planning and decision-making have been made \cite{kelly2003reactive,zhang2022rethinking}. However, challenges in interpretability and data limitations remain \cite{gohel2021explainable,arrieta2020explainable,singh2023recent,atakishiyev2021explainable,lin2024adaptsfl,lin2023fedsn}. To address these, recent research has integrated LLMs into autonomous systems for enhanced reasoning and interpretability \cite{fu2024drive,chen2023driving}. However, these approaches still grapple with translating LLM reasoning into practical driving maneuvers. LLMs have shown promise in autonomous driving with their advanced cognitive and reasoning capabilities \cite{radford2018improving,radford2019language,ouyang2022training}. Applications of LLMs range from fine-tuning pre-trained models \cite{liu2023mtd,chen2023driving,xuDriveGPT4InterpretableEndtoend2023} to innovative prompt engineering \cite{wenDiLuKnowledgeDrivenApproach2023,cui2023receive}. Despite significant advances, challenges like inference speed and limited real-world application still persist. Additionally, LLMs have great potential to improve adaptability in perception stages in autonomous driving systems \cite{radford2021learning,mao2023language,qian2023nuscenes}. Techniques like PromptTrack \cite{wu2023language} and HiLM-D \cite{ding2023hilm} showcase LLMs' effectiveness in detection and tracking, yet how to effectively handle real-world complexity remains a significant challenge. 


The advances of LLMs have shown great human-like abilities such as understanding, reasoning, and zero-shot learning. These abilities make LLMs a promising choice for autonomous driving (AD) across decision-making, scenario understanding, and planning. For example, Cui et al. \cite{cuiDriveYouSpeak2023} leveraged the language and reasoning abilities of LLMs into autonomous vehicles to realize an interactive design by human-machine dialogues. 
Wen \textit{et al.} \cite{wenDiLuKnowledgeDrivenApproach2023}  proposed DiLu, a LLM-based close-loop knowledge-driven approach that integrates memory and self-reflection modules to AD system to enhance the decision-making process. 
Mao \textit{et al.} proposed GPT-Driver \cite{maoGPTDriverLearningDrive2023}, a simple approach that leverages the pre-trained GPT-3.5 model to act like a motion planner for autonomous driving, and they also proposed Agent-Driver \cite{maoLanguageAgentAutonomous2023} to leverage LLMs to manipulate specific tools for AD. Sha \textit{et al.} \cite{shaLanguageMPCLargeLanguage2023} proposed LanguageMPC, a LLM-based model predictive control (MPC) framework that enables an autonomous vehicle to understand the natural language commands and execute the corresponding control commands.
However, the aforementioned works did not investigate the collaborative decision making for AD and lack comprehensive design of the framework. 
In this paper, we further exploit LLMs in high-level decision-making for autonomous driving and we carefully design reinforcement reflection and communication module to enhance the reliability and enable the collaborative decision-making of the framework, thereby advancing the field towards practical, LLM-driven collaborative driving systems.

In Table \ref{tab:comparison_capabilities}, we compare the capabilities of \textsc{AgentsCoDriver} and other approaches.
Compared with other LLM-based methods for AD that offer different functions for autonomous driving. Our \textsc{AgentsCoDriver} encompasses a wide range of abilities to handle complex driving tasks. By incorporating \textit{cognitive memory, reinforcement reflection, multi-vehicle collaborative driving,} and \textit{inter-vehicle communications}, we will elaborate the details in the following sections.

\begin{table*}[t] 
    \centering
    \renewcommand\arraystretch{1.2}

    \caption{\textbf{Comparison of Capabilities for \textsc{AgentsCoDriver} and Other Approachs.}  `\textcolor{cyan}{\Checkmark}' indicates the presence of a specific ability in the corresponding framework, `\XSolidBrush' means the absence.}\vspace{2mm}

    \resizebox{1\textwidth}{!}{
    \begin{tabular}{c|ccccc}
        \Xhline{1.2pt} \rowcolor[HTML]{EFEFEF}
    Framework Capablity         & Agent-Driver \cite{maoLanguageAgentAutonomous2023} & DiLu \cite{wenDiLuKnowledgeDrivenApproach2023} & GPT-Driver \cite{maoGPTDriverLearningDrive2023}& LanguageMPC \cite{shaLanguageMPCLargeLanguage2023}& \textsc{AgentsCoDriver} \\ \Xhline{1.2pt}
     
    Cognitive Memory            &\textcolor{cyan}{\Checkmark}               &\textcolor{cyan}{\Checkmark}      &\XSolidBrush            & \XSolidBrush                   & \textcolor{cyan}{\Checkmark}              \\ \rowcolor[HTML]{EFEFEF}
    Self-Reflection             &\textcolor{cyan}{\Checkmark}               &\textcolor{cyan}{\Checkmark}      &\XSolidBrush            &  \XSolidBrush                  &\textcolor{cyan}{\Checkmark}                 \\ 
    Reinforcement Reflection             &\XSolidBrush              &\XSolidBrush      &\XSolidBrush            &  \XSolidBrush                  &\textcolor{cyan}{\Checkmark}                 \\\rowcolor[HTML]{EFEFEF} 
    
    Collaborative Driving &\XSolidBrush             &\XSolidBrush      &\XSolidBrush            & \XSolidBrush                   &\textcolor{cyan}{\Checkmark}                 \\
    Inter-Vehicle Communication &\XSolidBrush              &\XSolidBrush      &   \XSolidBrush         &      \XSolidBrush              &\textcolor{cyan}{\Checkmark}                 \\\rowcolor[HTML]{EFEFEF} 
    
    Lifelong Learning           &\XSolidBrush              &  \textcolor{cyan}{\Checkmark}    &  \XSolidBrush          &     \XSolidBrush               &\textcolor{cyan}{\Checkmark}                 \\ \Xhline{1.2pt}
    \end{tabular}}
    \label{tab:comparison_capabilities}

\end{table*}

\subsection{Multi-Agent Collaborative Driving}

Recently, intensive works have explored the LLM-based multi-agent collaboration and communication \cite{huAdaptiveCommunicationsCollaborative2023,huFullsceneDomainGeneralization2023,hu2024collaborative,palanisamyMultiAgentConnectedAutonomous2020,shalev-shwartzSafeMultiAgentReinforcement2016}. Zhang \textit{et al.} \cite{zhangBuildingCooperativeEmbodied2023a} proposed a framework that leverages LLMs to build collaborative embodied agents, which can communicate with each other to collaboratively achieve a design goal in various embodied environments. Hong \textit{et al.} \cite{hongMetaGPTMetaProgramming2023} designed MetaGPT, which is a meta-programming framework that integrates human work flows into LLM-based multi-agent collaboration. Li \textit{et al.} \cite{liCAMELCommunicativeAgents2023} developed CAMEL, a role-playing framework that allows different agents to communicate with each other to facilitate autonomous collaboration. As for connected and autonomous driving (CAD), several  works focus on collaborative perception \cite{chenVehicleServiceVaaS2023,fang2024pacp,zhouAlldayvehichledetection2023,zhouPedestrianCrossing2024}, which proactively leverage capability-empowered vehicles to enhance perception by sharing data among agents. In addition, promising datasets like V2X-SIM \cite{li2022v2x}, OPV2V \cite{xu2022opv2v}, and DAIR-V2X  \cite{yu2022dair} were collected for this purpose.  Several approaches, such as F-Cooper \cite{chen2019f} and V2VNet \cite{wang2020v2vnet}, were proposed to leverage specific features or graph neural networks to perform more efficient data integration. Some innovative  methods like DiscoNet \cite{mehr2019disconet} and Where2comm \cite{hu2022where2comm} attempted to optimize communications for better cooperation and/or take better advantage of early or intermediate data fusion. In this paper, we focus on leveraging LLMs to comprehensively facilitate end-to-end collaborative decision-making to boost the performance of safe driving. 

\begin{figure*}[t]
    \centering
    \includegraphics[width=0.9\textwidth]{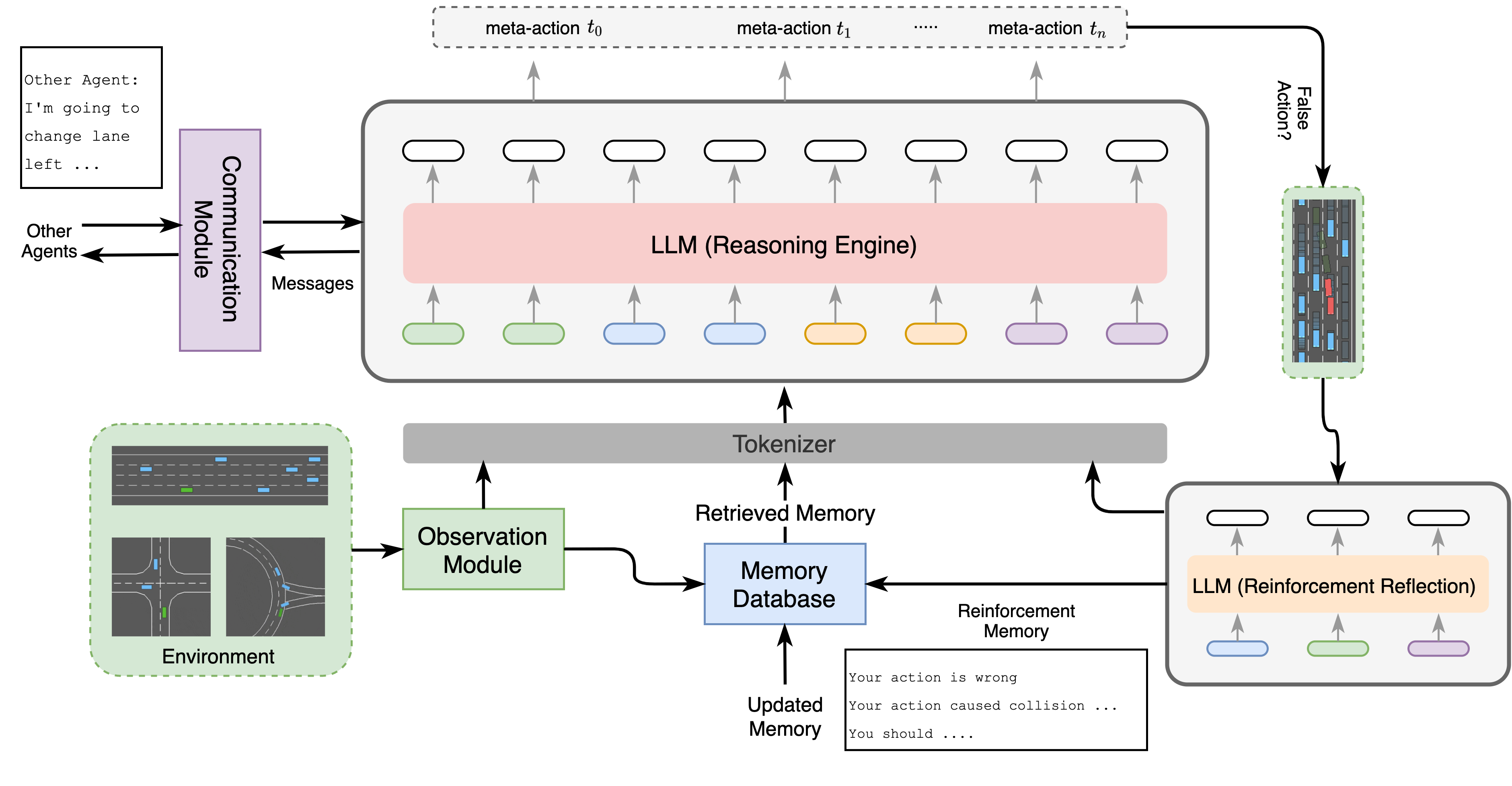}
    \caption{\textbf{Overall Architecture of \textsc{AgentsCoDriver}.} The architecture of \textsc{AgentsCoDriver} consists of five modules: \textit{observation module, reasoning engine, memory module, reinforcement reflection module,} and \textit{communication module}. The \textit{reasoning engine, communication module,} and \textit{reinforcement reflection module} leverage LLMs to generate messages and final decisions. }
    \label{fig:overall_architecture}
\end{figure*}


\section{System Formulation}
\label{sec:system_formulation}

The multi-vehicle collaborative driving can be formulated as a Decentralized Partially Observable Markov Decision Process (D-POMDP) with inter-agent communications \cite{bhallaDeepMultiAgent2020}. The D-POMDP mathematical model can be defined as follows.
\begin{equation}
    \langle A, S, \{A_i\}, \{O_i\}, Z, G \rangle
\end{equation}
where \( A \) denotes the set of agents and \( S \) represents the finite state space of the environment. Each agent \( i \) in \( A \) is associated with its own action space \( A_i \) and observation space \( O_i \), reflecting the decentralized nature of the decision-making process. Complementing these, the observation function is defined as follows. 
\begin{equation}
    Z: S \times A_1 \times ... \times A_n \rightarrow \Pi(O_1 \times ... \times O_n)
\end{equation}
which defines the probability distribution of the possible observations received by each agent, given the current state and actions. 
The crux of solving a D-POMDP involves devising a set of policies \( \{\pi_1, \pi_2, ..., \pi_n\} \), one for each agent, which map sequences of observations to actions, with the aim to realize the goal $G$.

In multi-vehicle collaborative driving, connected and autonomous vehicles (CAVs) collaborate with each other to improve their driving performance. CAV $i\in A$ takes action $a_i\in A_i$ to change lane, accelerate, overtake another CAV, and communicate with each other in a partially observable environment with the observation $o_i\in O_i$ at the current time instant $t$ and the messages from other CAVs to collaboratively realize a goal $g_i\in G$ with the policy $\pi_i$.

\section{Method: \textsc{AgentsCoDriver}}
\label{sec:method}

In this section, we propose \textsc{AgentsCoDriver}, an LLM-based multi-vehicle collaborative driving framework. We first introduce the overall architecture of \textsc{AgentsCoDriver} (Sec. \ref{sec:overall_architecture}), then we present the details of each module, including 1) observation module, 2) reasoning engine module, 3) memory module, 4) iterative reinforcement reflection module, and 5) communication protocol module (Sec. \ref{sec:observation}, \ref{sec:reasoning_engine}, \ref{sec:memory}, \ref{sec:reflection}, \ref{sec:communication_protocol}, respectively)

\subsection{Overall Architecture}
\label{sec:overall_architecture}

Fig. \ref{fig:overall_architecture} shows \textsc{AgentsCoDriver}, our multi-vehicle, closed-loop, and lifelong learning collaborative driving framework. 

At each step, the observation module first perceives the surrounding environments and extract necessary information. This information will be turned into a pre-defined structured description, which will be encoded into embeddings and used to recall the top-K related memory from the memory module, After that, the description, related memory, and the messages received from other agents (if any) will be combined into a prompt and fed to the reasoning engine. The reasoning engine performs multi-round reasoning based on the inputs and generate the final decision. Finally, the decision will be decoded into specific meta actions and executed at the ego CAV for safe driving. In addition, after generating a decision, the communication module will determine whether to communicate with other agents and what to communicate for. If the communication module determines to communicate with other agents, then its agent will communicate with other agents to exchange information. Finally, the evaluator and reflector will generate the reward score and verbal reinforcement analysis result for the agent's decision, respectively. The reward score and verbal reinforcement analysis result will be stored in the memory module.

\subsection{Observation Module}
\label{sec:observation}

In order to enable agents to collaborate, it is important for CAVs to perceive their surrounding environments and extract necessary information for downstream higher-order task reasoning, and so we develop an observation module for an agent to encode its  surrounding scenario and extract its useful high-level information such as the number of lanes and the location and velocity of its surrounding vehicles. These observations will then be fed into the agent's reasoning engine for analysis and make decisions. They are also used to recall the related memory from the memory module.

\subsection{Reasoning Engine}
\label{sec:reasoning_engine}

Reasoning is the fundamental and vital ability of humans, which is significant for the human to make daily and complex decisions. Traditional data-driven methods directly use the perception information (\textit{e.g.}, the object detection results and the semantic segmentation results) to make the final driving decisions (\textit{e.g.}, turning left or right, acceleration, and deceleration), which lacks interpretability and cannot handle complex scenarios and long-tailed cases. Inspired by the human's reasoning ability, we propose a reasoning engine for a CAV's agent, which consists of three steps: 1) prompt generation, 2) reasoning process, and 3) motion planning. 
The reasoning process is summarized in Algorithm \ref{alg:reasoning_engine}.

\begin{figure}[t]
    \centering
    \includegraphics[width=0.6\linewidth]{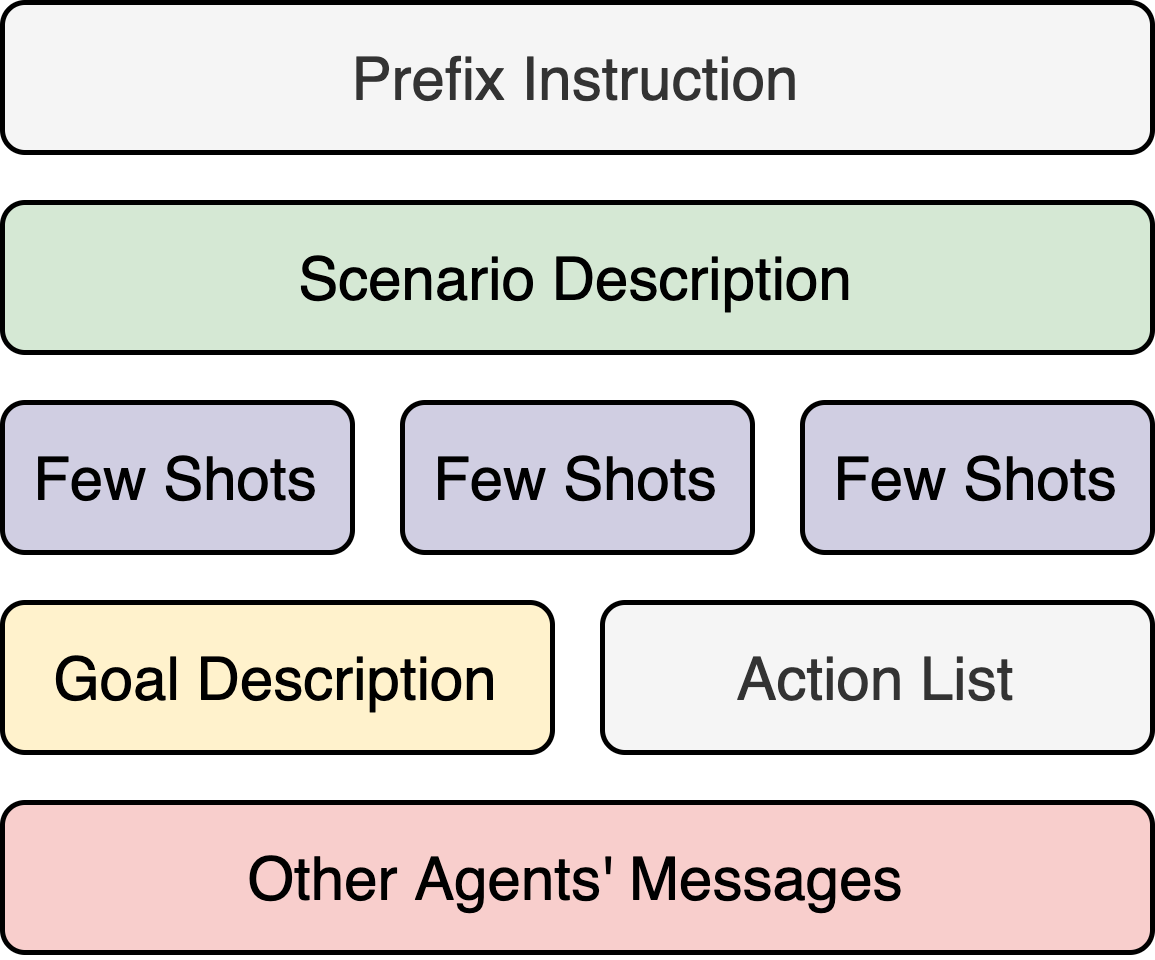}
    \caption{Prompt design of the reasoning engine.}
    \label{fig:reason_prompt}
\end{figure}

\textbf{Prompt Generation.} The prompt is divided into several parts: 1) prefix instruction, 2) scenario description, 3) few-shot experience, 4) goal description, 5) action list, and 6) the messages received from other agents (if any), as shown in Fig. \ref{fig:reason_prompt}.
Before generating a prompt, an agent should first leverage observation module described in Sec. \ref{sec:observation} to monitor its surroundings and clarify the static road details and dynamic vehicle information. After observation and obtaining the description of the environments, the description will be encoded into embeddings, which  will be used to recall the top-K related memory from the memory module, such as  experience memory used as few-shot experience. Then, these components will be combined into a pre-defined structured template and fed to the reasoning engine.

\textbf{Reasoning Process.}  With all the information provided by the previous modules, an agent needs to synthesize and reason over the current state, the messages from others, the scenario, the past experiences, the goals, and the commonsense knowledge to come up with a decision on what to do next. A strong reasoning engine is significantly needed to enable an agent to make effective correct decisions. In order to facilitate the reasoning ability, we directly harness LLMs as the reasoning engine with carefully designed prompts stated above to reason over all the information and generate the final decision. In addition, we also utilize the novel chain-of-thought prompting technique introduced in \cite{weiChainofThoughtPromptingElicits} to make LLMs decompose the complex problems into a series of sub-problems and solve them step by step by generating a series of intermediate outputs before making the final decisions. This strategy is employed because of the inherent complexity and variability in collaborative driving scenarios, which could result in inaccurate or hallucinations outcomes if LLMs were to directly generate decision results.

\textbf{Motion Planning.} After completing the reasoning process, an agent generates the final decision, which is then decoded into specific meta actions (including, but not limited to, idle, changing lane left, changing lane right, acceleration, deceleration, turning left, and turning right) to be executed by the agent, leading to the corresponding state transition in the environment. Then, the agent start to observe the environment again and repeat the above process. 

\subsection{Memory Module}
\label{sec:memory}

Memory is very important for a human. When a person drives a car, he or she will use the commonsense, such as complying with the traffic rules and recalling the past experiences to make decisions. In order to instill this ability into an agent, we propose a memory module for the agent, which consists of three parts: commonsense memory, experience memory, and reflection memory stored by structured texts. 
The commonsense memory contains the commonsense knowledge of driving, such as traffic rules. The experience memory contains the past driving scenarios and corresponding decisions. The reflection memory contains the feedback from the reflector module. The agent can retrieve the related memory from the memory module to make it available for decision-making.

\textbf{Commonsense Memory.} It consists of the basic but important knowledge that a driver should have, such as the basic traffic rules/regulations, basic driving rules, and the matters demanding attention. It is very important because it provides the basic knowledge for an agent to make driving decision. 

\textbf{Experience Memory.} It consists of the past driving scenarios, each of which contains the observation of the environment, the corresponding action, and the reasoning process. By retrieving and referencing experiences that are most similar, an agent's ability to make better-informed and resilient decisions can be improved.

\textbf{Reflection Memory.} It consists of the lessons learned and feedbacks from the \textit{reflector module} proposed in Sec. \ref{sec:reflection}, which helps an agent to learn from its historical mistakes and improve its future behavior. Specifically, it contains 1) the scenario description, 2) the effective reasoning processes, 3) the sound  decisions, and 4) the lessons learned.

\textbf{Memory Recall Method.} When people situate in a certain scenario or get  certain information, they can always recall similar memory to help them handle the current issue with similar decision. In order to realize human-like driving, a driving agent should also have this kind of ability. Inspired by the vector databases, we first encode the scenario description into embeddings and then we split the memory into items and encode each item into an embedding based on similarity measure. In this paper, we  calculate the similarity between a scenario description and  a scenario description part of each item in the embedding space, and return the top-K similar memory items.

In order to facilitate an agent to build its memory module, we first initialize the memory manually. For example, we select some scenarios and manually write the corresponding decisions and reasoning processes to the experience memory. In addition, we also select some mistaken scenarios and decisions and write the corresponding lessons and feedbacks to the reflection memory. At each iteration, the agent can retrieve the related memory from the memory module to make decisions. Finally, the agent, evaluator and reflector will store the related information into the memory module.

\begin{algorithm}[t]
    \caption{Reasoning Engine for Decision Making}
    \label{alg:reasoning_engine}
    \begin{algorithmic}[1]
    
    \Procedure{ReasoningEngine}{}
        \While{agent is operational}
            \State $P_i \gets PromptGeneration()$
            \State $P_i\_Tokens = Tokenizer(P_i)$
            \State $D_i \gets ReasoningProcess(P_i)$
            \State $MotionPlanning(D_i)$
        \EndWhile
    \EndProcedure
    
    \Procedure{PromptGeneration}{}
        \State \textbf{Input}: Lane number $L_n$, current lane of ego vehicle $L_c$, velocity $v$, position $(x, y)$; Surrounding vehicles $\{veh_1, \cdots, veh_i\}$ and corresponding velocity $v_i$ and position $(x_i, y_i)$; Messages $M$ from other agents.
        \State $Dec\gets Descriptor\{L_n, L_c, v, v_i, (x,y), (x_i, y_i), \cdots \}$
        \State $Scenario\_Tokens\gets Tokenizer(Dec)$
        \State Retrieve TopK related memories items: $MI_{1\sim K} \gets Memory\_Retriver(K, Scenario\_Tokens)$
        \State $P\gets Prompt\_Generator(Dec, M, MI_{1\sim K})$
        \State \Return Prompt $P$
    \EndProcedure
    
    \Procedure{ReasoningProcess}{$P$}
        \State Initialize LLM as $LLM(\cdot)$
        \State Split the task $T$ into sub-tasks $\{T_1, \cdots, T_n\}$
        \For {$T_i$ in $\{T_1, \cdots, T_n\}$}
            \State $O_i \gets LLM(T_i+P)$
            \State $P \gets P + O_i$
        \EndFor
        \State Generate final decision $D$
        \State \Return Decision $D$
    \EndProcedure
    
    \Procedure{MotionPlanning}{$Decision$}
        \State Decode $Decision$ into specific actions, such as {idle,  lane left,  lane right, acceleration, deceleration}, etc.
        \State Execute actions leading to state transition
    \EndProcedure
    
    \end{algorithmic}
    \end{algorithm}

\subsection{Reinforcement Reflection}
\label{sec:reflection}

\begin{figure}[t]
    \centering
    \includegraphics[width=1\linewidth]{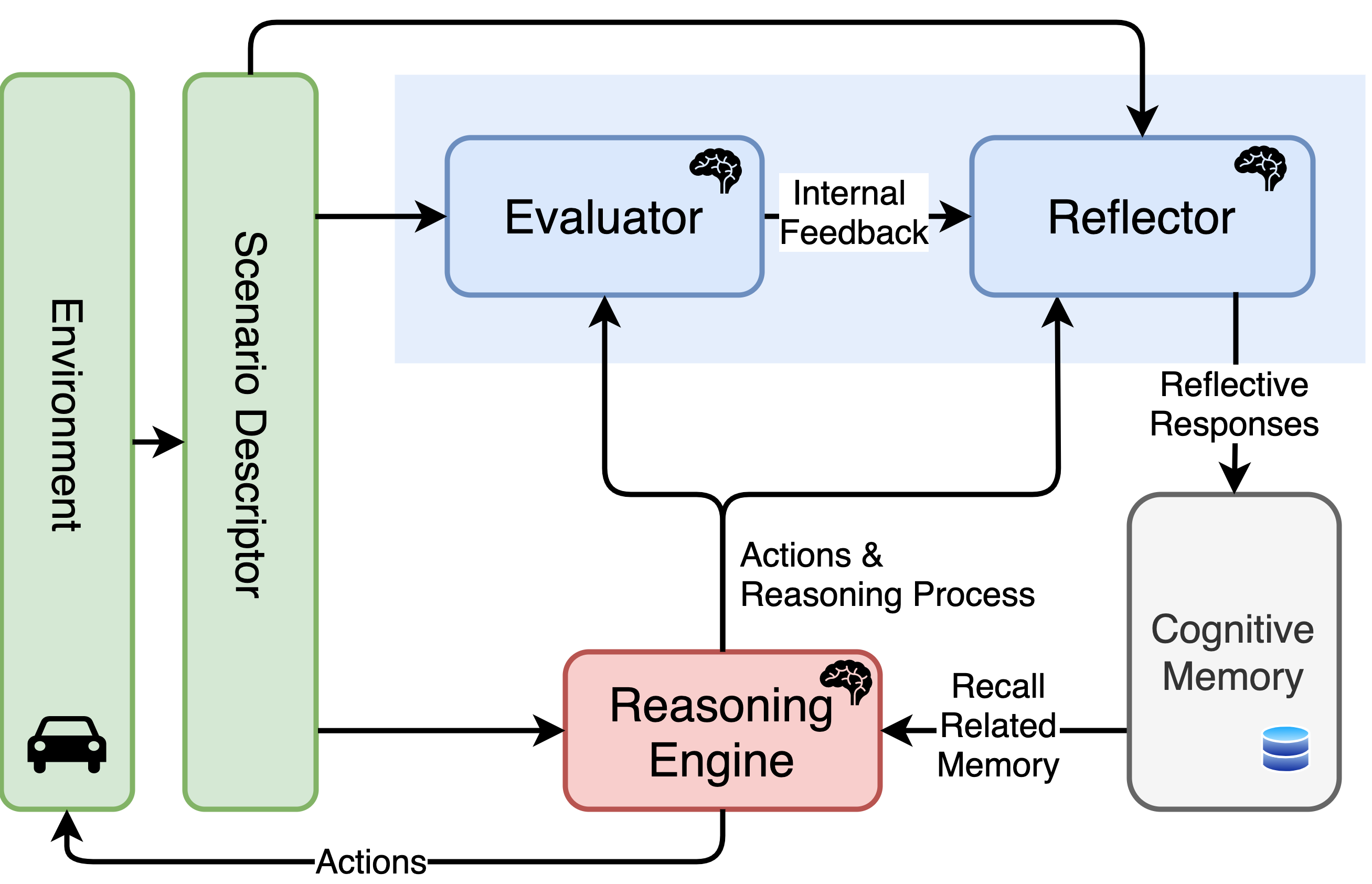}
    \caption{Reinforcement Reflection Module. 
    }
    \label{fig:iterative_reflection}
\end{figure}

If a person wants to become a specialist in a certain field, he or she must learn from his/her past experiences, which means that he or she must have the ability to introspect on his/her past mistakes and analyze the reasons behind them. For an agent who drives a car, it is also vital to have such an ability to reflect on itself to maintain the correct operations and safe driving. 
Based on these observations, we propose reinforcement reflection, which has two modules: an \textit{evaluator} and a \textit{reflector}. The evaluator, denoted as $E$, like a judge to score the output of an agent, and the reflector, denoted as $R$, can reflect on the agent's behavior and generate verbal reinforcement analysis result to improve the agent's actions. Different from the traditional reinforcement learning, this reinforcement reflection module generates the rewards by the evaluator.
We introduce these modules in detail in the following sections. The overall architecture of the reinforcement reflection module is shown in Fig. \ref{fig:iterative_reflection}. 

\textbf{Evaluator.} In the reinforcement reflection module, the evaluator plays an important role in assessing the quality of an agent's decisions. The evaluator takes the environment observation and the output (including the decision and reasoning process) of the agent to generate the reward score that represents whether the output is good or not. To assess the output of decision-making processes, we utilize a different instantiation of an LLM and specify its role as an evaluator, which generates the reward score for decision-making tasks.

\textbf{Reflector.} In order to build the reflector, we instantiate another LLM. The reflector is the core of a reinforcement reflection, which can generate verbal reinforcement feedbacks for an agent's output to help the agent learn from its historical mistakes. Specifically, based on the inputs of the reflector, namely, the reward signal from the evaluator $E$, the decision and reasoning process of the reasoning agent, and the scenario description, the reflector $R$ generates the feedbacks. Such feedbacks are very detailed about the reasons behind the incorrect actions of the agent, such as the lessons that the agent should learn, the suggestions for the agent to improve its future behavior, and the correct decision that the agent should output. All detailed feedbacks are stored in the agent's memory. 

\subsection{Communication Module}\label{sec:communication_protocol}

It is vital for collaborative agents to communicate with each other effectively. 
By communicating with each other, the observation range of the agent will be expanded. Consider two agents' observations $o_1$ and $o_2$, respectively, if the two agents exchange the observation information with each other, then the observation of the two agents will be expanded to $o_1 \cup  o_2$. In addition, the communication is also vital for the agents to negotiate with each other and make a better decision. For example, consider that an agent is driving a car and wants to overtake the front vehicle. If the agent communicates with the front vehicle, the agent of the front vehicle learns the intention of the vehicle behind, and then it can make better decisions to avoid potential collision. In order to realize effective communications, it is important to know when to communicate and what to communicate for. 

\textbf{When to Communicate.} Consider a scenario where several vehicles drive on the same road, an ego vehicle wants to execute certain actions (\textit{e.g.}, overtaking, turning left, and turning right), it is not always necessary for the ego vehicle to communicate with other vehicles, especially if they are far away. In such cases, the actions of the ego vehicle will not affect them, nor will there be any trajectory conflicts. Therefore, it is necessary to determine when to communicate. In order to do so, we directly leverage the agent itself to determine whether to communicate at each step.

\textbf{What to Communicate for.} In order to deal with this issue, we instantiate an LLM as the message generator. We integrate this generator as a tool that the agent can use. When the agent thinks it is necessary to communicate with others, it will call this tool and feed the current encoded scenario, the messages from the other agents (if any), and the previous action into the message generator, and then generate the messages to communicate with other agents. In order to facilitate communications, we design a structured prompt template, which consists of \textit{prefix instruction, action history, dialogue history,} and \textit{scenario description.} 

\begin{table*}[t]
    \centering
    \caption{\textbf{Main Results.} We evaluate \textsc{AgentsCoDriver} in two scenarios: \textit{Highway} and \textit{Intersection}. In each scenario, we evaluate the successful rate (SS) under two different settings: single-vehicle setting and two-vehicle setting. In each setting, we vary the number of memory items (shots).}
    \vspace{2mm}
    \renewcommand\arraystretch{1.1}
    \resizebox{0.9\textwidth}{!}{
    \begin{tabular}{c|c|c|cccccc}
    \Xhline{1.2pt}\rowcolor[HTML]{EFEFEF} 
    { Scenario}                    & No. of agents          & No. of shots & $\text{SS}_{min}$ &$\text{SS}_{Q1}$ & $\text{SS}_{median}$ & $\text{SS}_{Q3}$ & $\text{SS}_{max}$ & $\text{SS}_{mean}$ \\ \Xhline{1.2pt} 
    \multirow{8}{*}{\textbf{Highway}}   & \multirow{4}{*}{2 agents} & 0 shot          &2.0     &2.0    &4.0       &7.0    &12.0     &4.2      \\
                                &                           & 1 shot          &2.0     &3.0    &5.0        & 6.0   & 14.0    &5.2      \\
                                &                           & 3 shots         &4.0     &5.0    &8.0       &17.0    &30.0     &11.0      \\
                                &                           & 5 shots         &4.0     &6.0    & 8.0       &20.0    & 30.0    & 13.5     \\ \cline{2-9} 
                                & \multirow{4}{*}{1 agents} & 0 shot          &2.0     &3.0    &3.5        &6.0    &20.0     &5.6      \\
                                &                           & 1 shot          &3.0     &3.0    & 4.0       &5.0    &23.0     & 6.4     \\
                                &                           & 3 shots         &4.0     &5.0    &5.5        &10.0    &30.0     & 9.1     \\
                                &                           & 5 shots         &4.0     &5.0    & 16.5       & 30.0    & 30.0    & 17.3     \\ \Xhline{1.2pt}
    \multirow{8}{*}{\textbf{Intersection}} 
    & \multirow{4}{*}{2 agents} & 0 shot       &4.0  &6.0  &6.5  &10.0  &20.0  &8.2   \\
    &               & 1 shot       &5.0  &6.0  &7.5  &9.0   &30.0  &10.4  \\
    &               & 3 shots      &6.0  &8.0  &12.0 &21.0  &30.0  &15.6  \\
    &               & 5 shots      &4.0  &7.0  &15.0 &30.0  &30.0  &17.6  \\ \cline{2-9} 
    & \multirow{4}{*}{1 agents} & 0 shot  &5.0  &6.0  &7.0  &9.0   &10.0  &7.1   \\
    &                           & 1 shot       &5.0  &6.0  &10.0 &18.0  &28.0  &12.1  \\
    &                           & 3 shots      &3.0  &6.0  &8.5  &20.0  &30.0  &13.0  \\
    &                           & 5 shots      &5.0  &7.0  &25.0 &30.0  &30.0  &19.6
\\\Xhline{1.2pt}
    \end{tabular}
    }
    \label{tab:main_resultsss}

\end{table*}

\begin{figure}[t]
    \centering
    \includegraphics[width=0.8\linewidth]{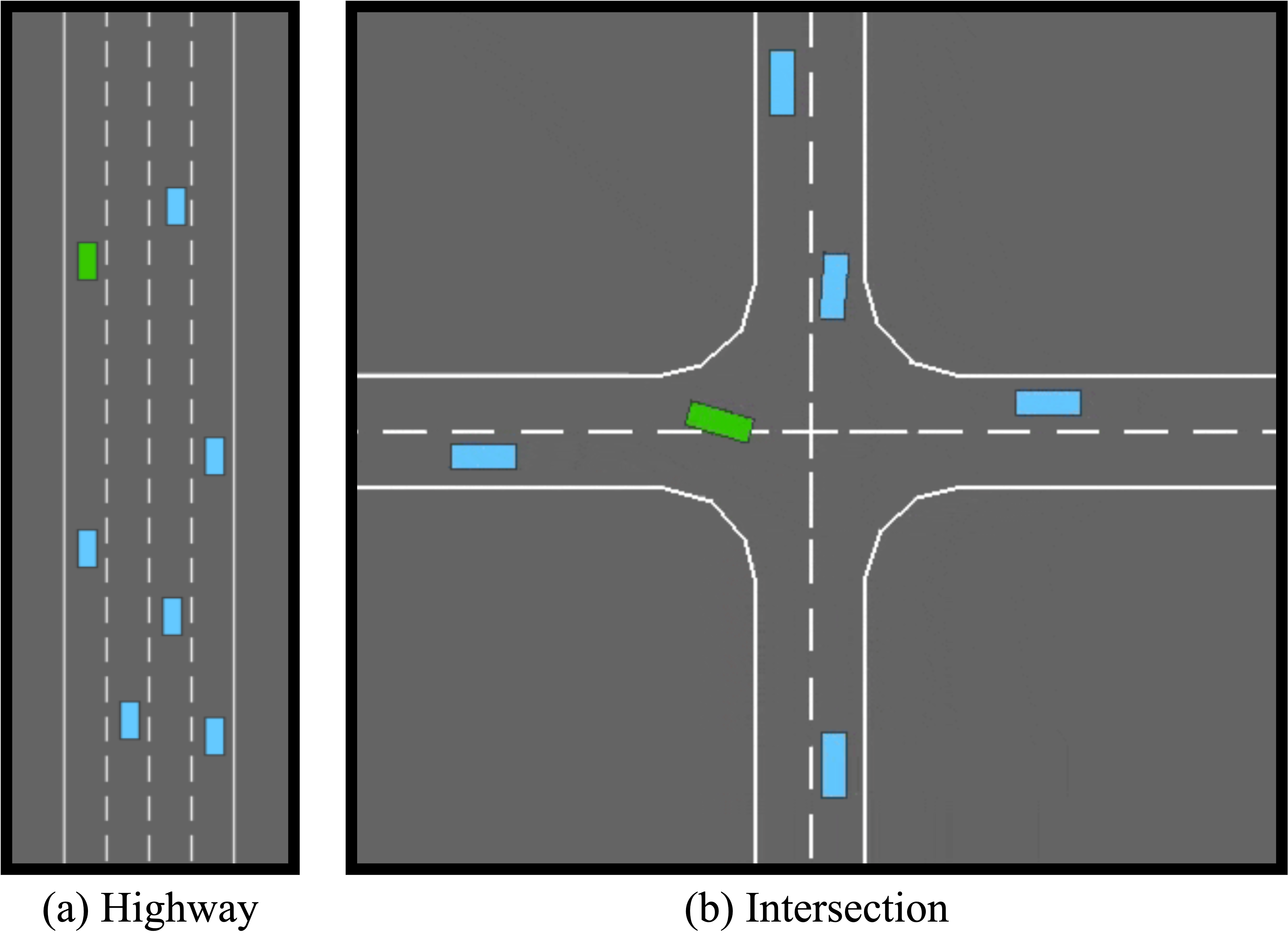}
    \caption{Visualization of two different scenarios. In \textit{Highway} scenario, the ego-vehicle (The \textcolor{ForestGreen}{green} one) is traveling on a highway with multiple lanes with various vehicles (The \textcolor{cyan}{blue} one). The goal of an agent is to achieve high velocity while ensuring it does not collide with nearby vehicles. In \textit{Intersection} scenario, the task of the ego vehicle is to pass a busy intersection safely without any collision. The ego vehicle can turn left or right, or go straight. }
    \label{fig:scenarios}
\end{figure}

\section{Experiments}
\label{sec:experiments}

\definecolor{myblue}{HTML}{6E94E7}

\begin{figure*}[t]
    \centering
    \subfloat[{Highway}]{
        \includegraphics[width=0.45\textwidth]{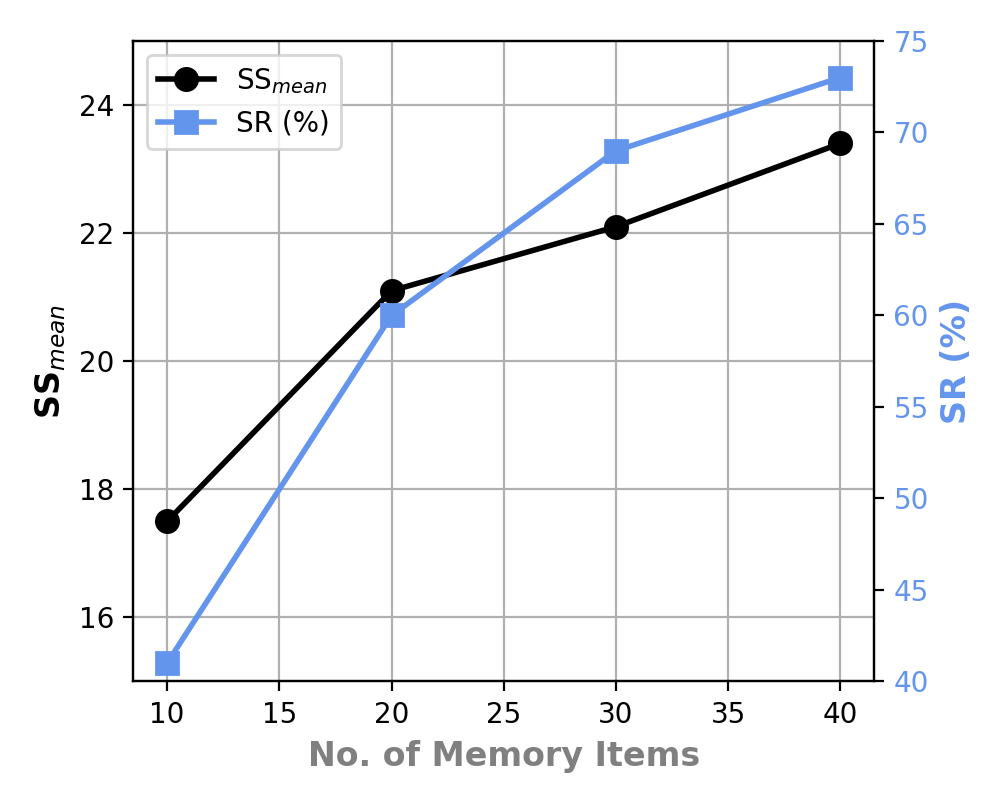}
        \label{fig:lifelong_highway}
    }
    \subfloat[Intersection]{
        \includegraphics[width=0.45\textwidth]{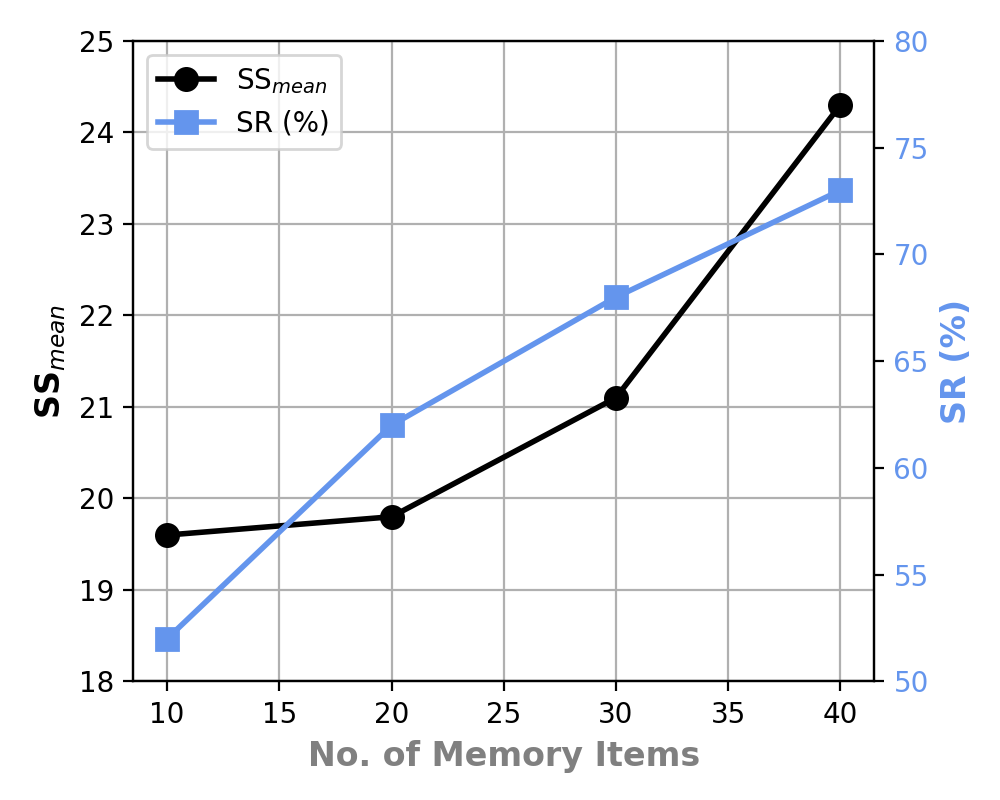}
        \label{fig:lifelong_intersection}
    }
    \caption{\textbf{Results of Lifelong Learning.} The above figure shows the results of lifelong learning where the \textbf{black} line represents the average successful steps (SS$_{mean}$), the \textcolor{myblue}{blue} line represents the successful rate (SR).}
    \label{fig:lifelong_combined}
\end{figure*}

\subsection{Experimental Setup}

\textbf{Simulation Environment.} In our experimental setup, we leverage widely recognized HighwayEnv \cite{highway-env} as our simulation environment, a popular platform in autonomous driving and tactical decision-making studies. This platform offers various driving models and effectively simulates interactions among multiple vehicles. 

\textbf{Implementation Details.} We build \textsc{AgentsCoDriver} by LangChain, which is a framework for developing LLM-based applications. We leverage GPT-3.5-turbo as our base LLM model, which is developed by OpenAI.


\textbf{Evaluation Metrics.} In order to evaluate the performance of the framework in HighwayEnv simulator, we adopt Success Rate (SR) and Success Step (SS) as our evaluation metrics. In our experiments, followed by \cite{wenDiLuKnowledgeDrivenApproach2023}, we define a successful completion time of 30 seconds with a decision-making frequency of 1Hz. If all the agents can drive without collisions for 30 times of decisions, which means SS is 30, we consider it as a successful episode. SR is the ratio of the number of successful episodes $S$ to the total number of episodes $S_{a}$, which can be formulated as: $${\text{SR}} = {S}/{S_{a}}\times 100\%$$
In addition, SS can be divided into $\text{SS}_{min}$, $\text{SS}_{Q1}$, $\text{SS}_{median}$, $\text{SS}_{Q3}$, $\text{SS}_{max}$, and $\text{SS}_{mean}$. In $N$ times experiments, $\text{SS}_{min}$ is the minimum value of SS, $\text{SS}_{Q1}$ is the first quartile of SS, $\text{SS}_{median}$ is the median of SS, $\text{SS}_{Q3}$ is the third quartile of SS, $\text{SS}_{max}$ is the maximum value of SS, and $\text{SS}_{mean}$ is the mean value of SS. $\text{SS}_{mean}$ can be formulated as: 
\vspace{-2mm}
\begin{equation}
    \text{SS}_{mean} = \frac{1}{N}\sum_{i=1}^{N}\text{SS}_{i}
    \vspace{-2mm}
\end{equation}
where $N$ is the repeated times of one episode. 


\subsection{Evaluation of \textsc{AgentsCoDriver}}

In this subsection, we mainly focus on evaluating the effectiveness of \textsc{AgentsCoDriver}, particularly on the reasoning process with or without cognitive memory. We evaluate \textsc{AgentsCoDriver} in two different scenarios in Highway simulator: \textit{Highway} and \textit{Intersection}, which are shown in Fig. \ref{fig:scenarios}. In each scenario, we initialize the number of memory items to 10, which are used under different few-shot settings. In each experiment, we set the number of shots as 0, 1, 3, or 5, which means in the process of reasoning, the agent will recall 0, 1, 3, or 5 memory items from the memory module. In addition, we conduct comparative experiments between single-vehicle setting and two-vehicle collaborative driving setting. Furthermore, each setting is repeated 10 times with different initialization seeds to obtain the final results, including SS$_{min}$, SS$_{Q1}$, SS$_{median}$, SS$_{Q3}$, SS$_{max}$, and SS$_{mean}$. The results are shown in Table \ref{tab:main_resultsss}.

We observe that with the increase of the number of shots, the performance significantly improves in both scenarios and under both single-vehicle setting and two-vehicle collaborative driving setting.
The average successful steps (SS$_{mean}$) increase by around 10 steps under all settings with the increase of the number of shots from 0 to 5. Furthermore, we observe that when the number of shots is 0, SS$_{mean}$ is very low and even does not exceed 10 steps. This phenomenon demonstrates that LLM cannot directly execute the driving tasks without any adaptation. At the same time, it also demonstrates the significance and importance of the cognitive memory module for the framework.

In addition, we find that SS$_{mean}$ under single-vehicle setting is higher than that under two-vehicle collaborative driving setting. For example, in Highway scenario, when the number of shots is 5, SS$_{mean}$ under single-vehicle setting is 17.3, while SS$_{mean}$ under two-vehicle collaborative driving setting is 13.5. After further analysis, we find it reasonable. We set the number of other vehicles (except the vehicles controlled by our framework) to 15, these vehicles are controlled by the simulator, and their behaviors do not result in collisions. In contrast, our framework is not perfect yet and the vehicles controlled by it has a non-negligible probability of collisions. Therefore, the more controlled vehicles, the higher the probability of collision with other vehicles. 
However, in Section \ref{sec:ablation_study}, we examine the impact of the multi-vehicle setting on the controlled ego vehicles, independent of the influence of other vehicles. The results demonstrate that a multi-vehicle setting with collaboration can enhance the SR of the framework.

\subsection{{Comparative Results}}

\begin{table}[t]
    \centering
    \renewcommand\arraystretch{1.3}

    \caption{Comparative experiments in HighwayEnv simulator. The three settings are conducted with 20 memory items with one, three, and five shots, respectively.}
    \begin{tabular}{c|cccccc}
    \Xhline{1.2pt}\rowcolor[HTML]{EFEFEF} 
    Methods& SS$_{min}$ & SS$_{Q1}$ & SS$_{median}$ & SS$_{Q3}$ & SS$_{max}$  \\ \Xhline{1.2pt}
    DiLu (Setting 1)&2.0&2.0&7.0&11.0&11.0\\\hline
    Ours (Setting 1)&\textbf{5.0}&\textbf{6.0}&\textbf{10.0}&\textbf{18.0}&\textbf{28.0}\\\Xhline{1.2pt}
    DiLu (Setting 2)&4.0&6.0&9.0&\textbf{29.0}&{30.0}\\\hline
    Ours (Setting 2)&\textbf{6.0}&\textbf{8.0}&\textbf{12.0}&21.0&\textbf{30.0}\\\Xhline{1.2pt}
    DiLu (Setting 3)&4.0&7.0&12.0&30.0&30.0\\\hline
    Ours (Setting 3)&\textbf{4.0}&\textbf{7.0}&\textbf{15.0}&\textbf{30.0}&\textbf{30.0}\\\Xhline{1.2pt}
    \end{tabular}
    \label{tab:comparative_results_highwayenv}
\end{table}



In this section, we conduct comparative experiments to evaluate the performance of \textsc{AgentsCoDriver} with other state-of-the-art methods. 
First, we conduct comparative experiments in HighwayEnv with DiLu \cite{wenDiLuKnowledgeDrivenApproach2023} in three settings with 20 memory shots. The first setting is 1 shot, the second setting is 3 shots, and the third setting is 5 shots. As shown in Table \ref{tab:comparative_results_highwayenv}, our method is superior to DiLu in all settings. For example, in the first setting, the SS$_{min}$ of our method is 5.0 while the SS$_{min}$ of DiLu is 2.0. In the second setting, the SS$_{median}$ of our method is 12.0 while the SS$_{median}$ of DiLu is 9.0. In the third setting, the SS$_{median}$ of our method is 15.0  while the SS$_{median}$ of DiLu is 12.0.


\subsection{Lifelong Learning}

Lifelong learning is an important ability for human-driver, which allows a  driver to continuously learn from his/her experiences and improve the future actions. 
In this section, we evaluate the lifelong learning ability of \textsc{AgentsCoDriver}. As shown in Fig. \ref{fig:lifelong_combined}, we evaluate it in two different scenarios: \textit{Highway} and \textit{Intersection}. In each scenario, we set the number of memory items to 10, 20, 30, and 40, respectively. 
Firstly, in Highway scenario shown in Fig. \ref{fig:lifelong_highway}, the  black line represents the average of successful steps ($\text{SS}_{mean}$), which shows a generally increasing trend as the number of memory items increases from 10 to 40. SS$_{mean}$ starts at just under 18 and rises to slightly over 23. The blue line represents the successful rate (SR), which also shows an increasing trend with the number of memory items. SR starts at just above 40\% and rises to just under 75\%. Secondly, in Intersection scenario shown in Fig. \ref{fig:lifelong_intersection},  $\text{SS}_{mean}$ and SR also show  increasing trend with the number of memory items. These results demonstrate that \textsc{AgentsCoDriver} can effectively improve the performance of lifelong learning in different scenarios.

\begin{table}[t]
    \renewcommand\arraystretch{1.3}

    \caption{Ablation study of the iterative reinforcement reflection. `w/o' means without reflection, `w' means with reflection.}
    \resizebox{1\linewidth}{!}{
    \begin{tabular}{c|cccccc}
    \Xhline{1.2pt}\rowcolor[HTML]{EFEFEF} 
    Methods& SS$_{min}$ & SS$_{Q1}$ & SS$_{median}$ & SS$_{Q3}$ & SS$_{max}$ & SS$_{mean}$ \\ \Xhline{1.2pt}
    w/o&2.0&3.0&5.0&10.0&30.0&9.5\\\hline
    w&4.0&7.0&14.0&28.0&30.0&17.5\\\Xhline{1.2pt}
    \end{tabular}}
    \label{tab:ablation_study_iterative_reflection}
\end{table}

\subsection{Ablation Study}

\label{sec:ablation_study}
\textbf{Effect of the reflection.} In order to evaluate the effect of the iterative reinforcement reflection module, we conduct an ablation study. We compare the performance of the framework with and without the iterative reinforcement reflection module. We consider the single-vehicle setting with the number of shots set to 5. First, we randomly sample 5 memory items without reflection as the 5 shots, and then we randomly sample 5 reflected memory items. In each setting, we repeat 10 times with different initialization seeds to obtain the final results.
The results are shown in Table \ref{tab:ablation_study_iterative_reflection}. We observe that  SS$_{mean}$ of the framework with reflection is higher than that of the framework without reflection. This phenomenon demonstrates that the reinforcement reflection module can effectively improve the performance of our framework.

\begin{figure}[t]
    \centering
    \includegraphics[width=0.8\linewidth]{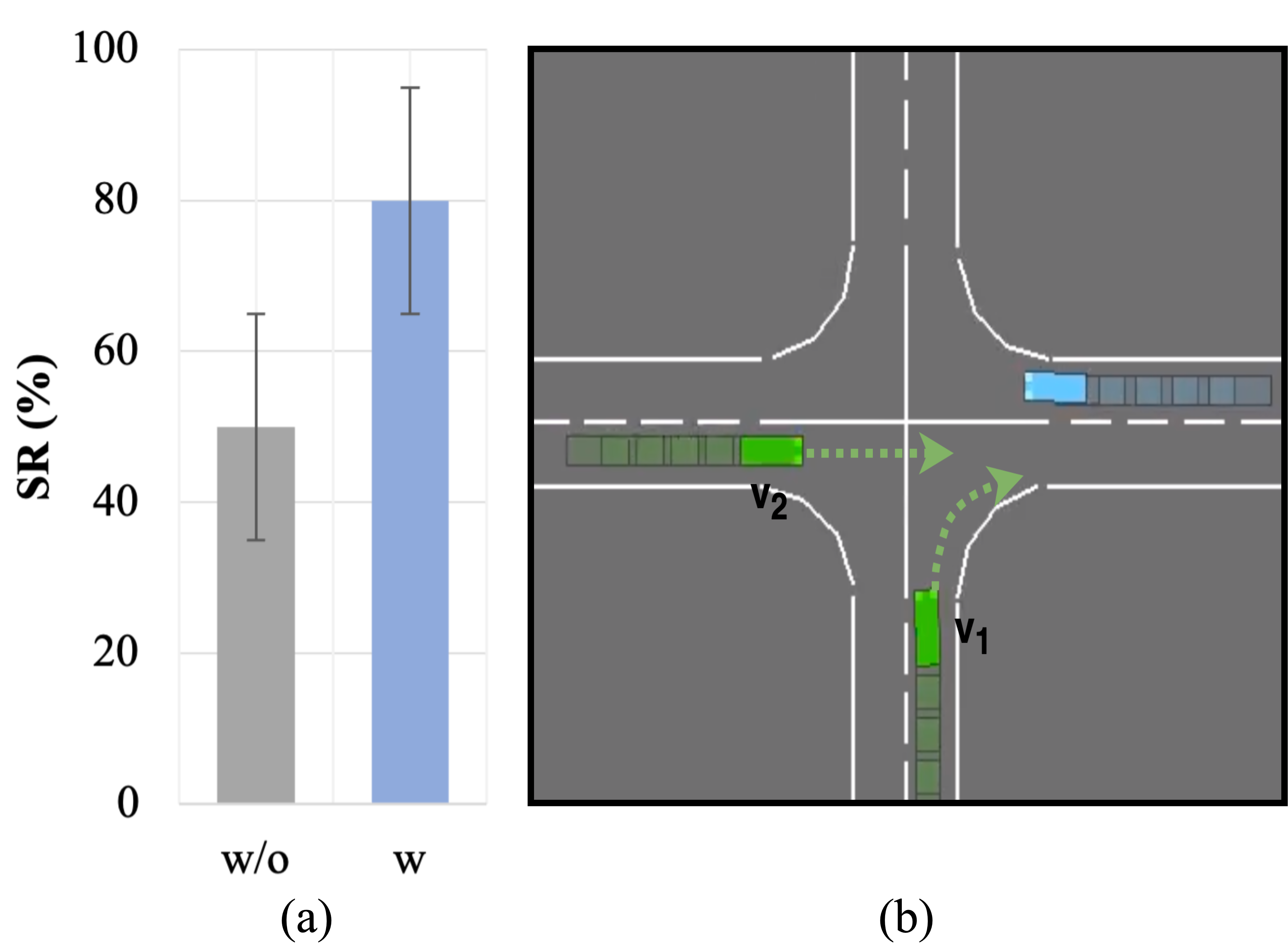}

    \caption{Ablation study of the communication module. Subfigure (a) shows the SR without and with the communication, where `w/o' means without communication, `w' means with communication. Subfigure (b) is the scenario of the multi-vehicle communication and collaboration.}
    \label{fig:comm_ablation}
\end{figure}

\begin{table}[t]
    \centering
    \renewcommand\arraystretch{1.3}

    \caption{Ablation study of the communication module. `w/o' means without communication or collaboration, `w' means with communication or collaboration.}
    \begin{tabular}{c|cccccc}
    \Xhline{1.2pt}\rowcolor[HTML]{EFEFEF} 
    NO. of Agents& SR (w/o) & SR (w) & Growth Ratio \\ \Xhline{1.2pt}
    3&45.0\%&75.0\%&66.7\%\\\hline
    5&40.0\%&70.0\%&75.0\%\\\Xhline{1.2pt}
    \end{tabular}
    \label{tab: more_vehicles}
\end{table}
\begin{figure}[t]
    \centering
    \includegraphics[width=0.9\linewidth]{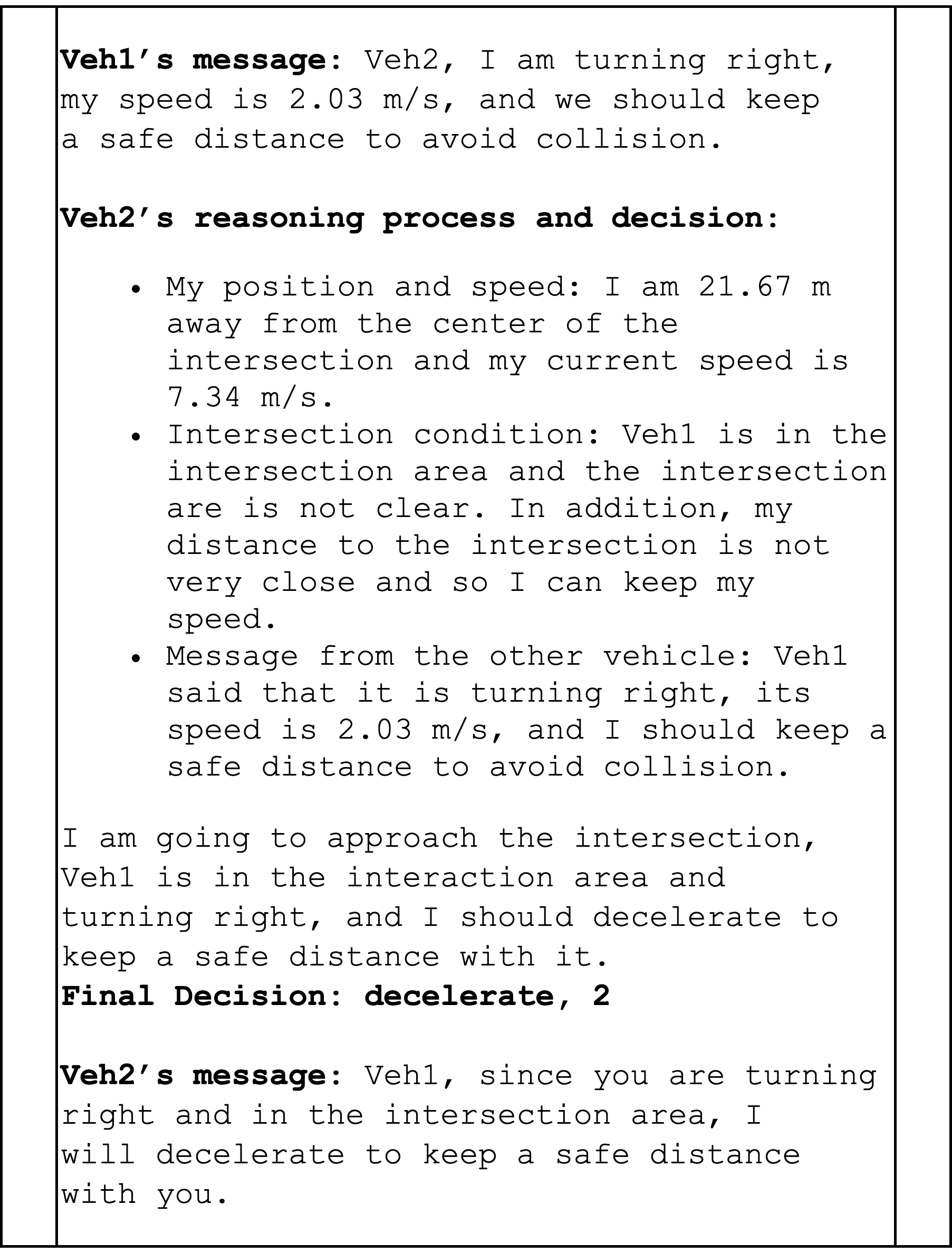}

    \caption{Reasoning process and communication messages between two vehicles.}
    \label{fig:case_study}
\end{figure}

\textbf{Effect of the communications.} In order to evaluate the effect of the communication module, we conduct an ablation study. We compare the performance of our framework with and without the communications. In order to reduce or eliminate the impact of the other vehicles (controlled by the simulator), we reduce the number of other vehicles to 1 so that it will not affect the controlled vehicles.
Consider the intersection scenario as shown in Fig. \ref{fig:comm_ablation}(b), two controlled vehicles are going to across the interaction. First vehicle $v_1$ approaches the intersection from south to north and plans to turn right. Second vehicle $v_2$ approaches the intersection from west to east and plans to go straight. In this scenario, $v_1$ and $v_2$ need to communicate and negotiate with each other to avoid collision and increase traffic efficiency. The results are shown in Fig. \ref{fig:comm_ablation}(a), it can be observed that SR of our framework with communications significantly improves the performance compared to the framework without communications, from around 50\% to 80\%.
This phenomenon demonstrates that the communication module can effectively improve the performance of our framework. 
In order to further elaborate the communication and decision process, we provide the reasoning process and communication messages between two vehicles as shown in Fig. \ref{fig:case_study}. We can see the two vehicles make decisions not only based on their own observations, but also based on the messages from the other vehicle.

In addition, we conduct more experiments with more than two vehicles. As shown in Table \ref{tab: more_vehicles}, we consider the intersection scenario with 3 and 5 controlled vehicles. We observe that the growth ratio of SR is 66.7\% and 75.0\%, respectively, which shows the effectiveness of the communication and collaboration among multiple vehicles.



\subsection{Qualitative Analysis}

\begin{figure*}[t]
    \centering
    \includegraphics[width=0.9\linewidth]{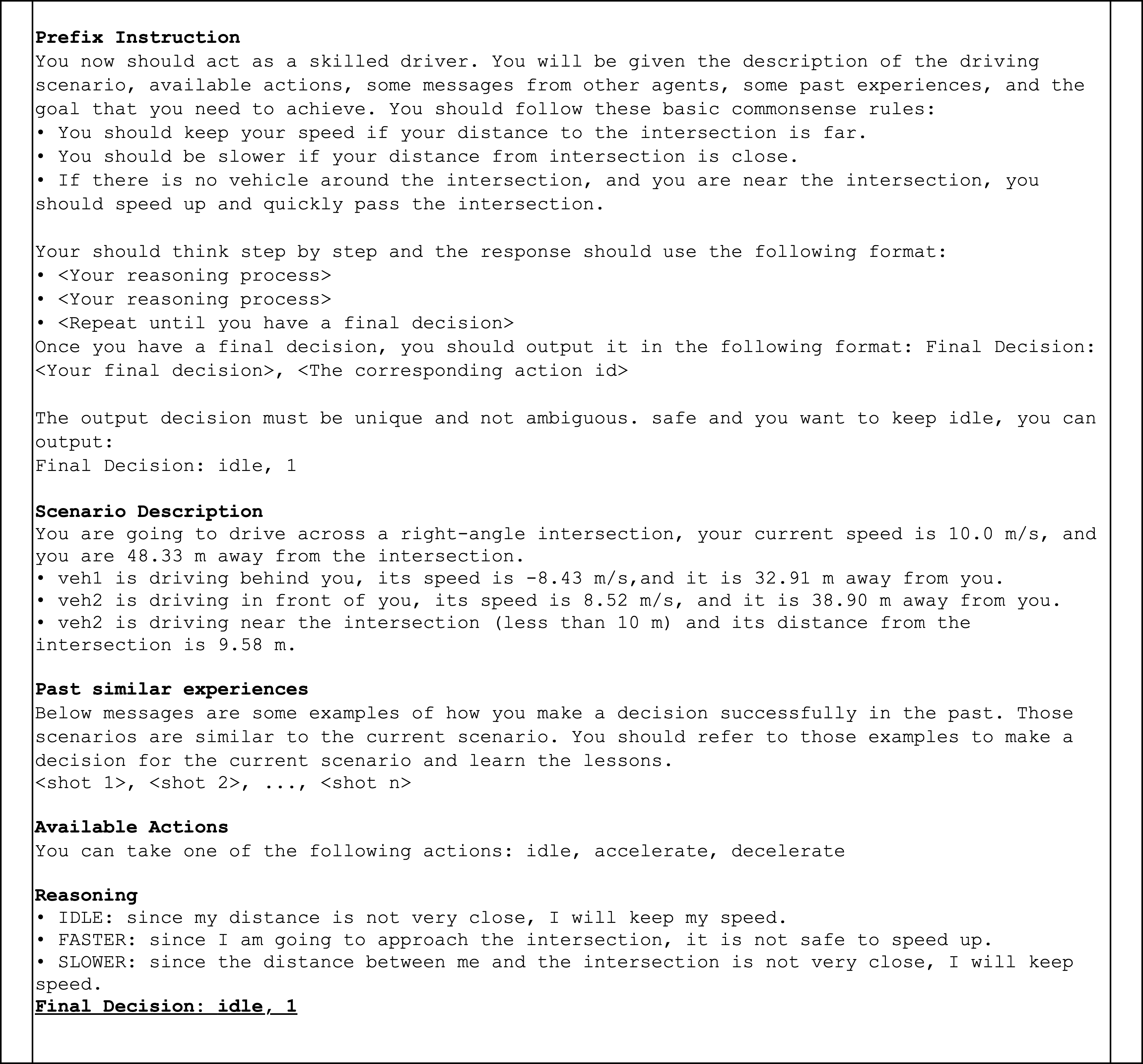}

    \caption{Example Reasoning Process of the Agent}
    \label{fig:qualitative_result}
\end{figure*}

In this section, we provide a qualitative analysis of the framework as shown in Fig. \ref{fig:qualitative_result}. \textsc{AgentsCoDriver} first obtain the related key environmental information, include the road condition, ego state, and other vehicle's state. Then the ego vehicle retrieves the related few shots from its own memory base. 
Then, the agent will reason based on the previous information and obtain the final decision. From the reason process, we observe that the agent successfully analyzed the current situation and made the correct decision.

\section{Limitations and Future Works}
\label{sec:limitations}

\textsc{AgentsCoDriver} is an effective framework for multi-vehicle collaborative driving and performs well under different experimental settings. However, it still has some limitations. First, like other LLM-based agents, it is not real-time and requires several seconds to generate the output. This limits the applicability in real-world connected and autonomous driving. In addition, \textsc{AgentsCoDriver} is text-based and lacks the ability to directly understand the visual information. Therefore, in our future research, we will extend \textsc{AgentsCoDriver} to multi-modal collaborative driving in real-world scenarios. In addition, we will also focus on the latency issue when applying LLMs to collaborative driving.

\section{Conclusion}
\label{sec:conclusion}

In this paper, we have proposed \textsc{AgentsCoDriver}, an LLM-powered multi-vehicle collaborative driving framework with lifelong learning, which allows different driving agents to communicate with each other and collaboratively drive in complex traffic scenarios. It features reasoning engine, cognitive memory,  reinforcement reflection, and communication module. The cognitive memory module and reflection module make the \textsc{AgentsCoDriver} capable of lifelong learning. 
Extensive experiments show the superiority of \textsc{AgentsCoDriver}. To the best of our knowledge, this is the first work to leverage LLMs to enable multi-vehicle collaborative driving.


\bibliographystyle{IEEEtran}

{\small
\bibliography{ref, ref2}}

\vfill

\end{document}